\pdfoutput=1
\documentclass[conference]{IEEEtran}
\IEEEoverridecommandlockouts
\usepackage{amsmath,amssymb,amsfonts}
\usepackage[flushleft]{threeparttable}
\usepackage{array,booktabs,makecell}
\usepackage{graphicx}
\usepackage{textcomp}
\usepackage{xcolor}
\usepackage{paralist}
\usepackage{flushend}
\usepackage{balance}
\usepackage{comment}
\usepackage[ruled,vlined]{algorithm2e}
\usepackage{relsize}
\usepackage[capitalize]{cleveref}
\usepackage[detect-all,per-mode=symbol]{siunitx}
\usepackage[protrusion=true,expansion=true,final,babel]{microtype} 
\usepackage{pgfplots}
\usepackage{caption}
\usepackage{subcaption}

\usepackage{times}
\pgfplotsset{compat=1.14}

\usepackage[subtle]{savetrees}
\usepackage[compact]{titlesec}
\usepgfplotslibrary{groupplots}
\usetikzlibrary{positioning}
\pgfplotsset{every axis legend/.append style={%
cells={anchor=west}}
}
\pgfplotsset{every axis/.append style={
                    label style={font=\footnotesize},
					tick label style={font=\footnotesize},
					legend style={font=\footnotesize}
                    }}
\usetikzlibrary{arrows}
\tikzset{>=stealth'}
\usepackage[style=ieee,maxcitenames=2,mincitenames=1,natbib=true]{biblatex}
\AtEveryBibitem{
 	\clearfield{url}
 	\clearfield{doi}
\ifentrytype{inproceedings}{
 	\clearlist{address}
 	\clearlist{publisher}
 	\clearname{editor}
 	\clearlist{organization}
 	\clearfield{pages}
 	\clearlist{location}
	\clearfield{volume}
	\clearfield{series}
 }{}
 }

\AtBeginBibliography{\small}

\title{A Graph Attention Learning Approach to \\ Antenna Tilt Optimization}
\author{\IEEEauthorblockN{
Yifei Jin\IEEEauthorrefmark{1}\IEEEauthorrefmark{2}\textsuperscript{\textsection},
Filippo Vannella\IEEEauthorrefmark{1}\IEEEauthorrefmark{2}\textsuperscript{\textsection}, Maxime Bouton\IEEEauthorrefmark{2}, Jaeseong Jeong\IEEEauthorrefmark{2}, Ezeddin Al Hakim\IEEEauthorrefmark{2} \\
\IEEEauthorrefmark{1}KTH Royal Institute of Technology, Stockholm, Sweden\\
\IEEEauthorrefmark{2}Ericsson Research, Stockholm, Sweden\\
Email: \{yifeij, vannella\}@kth.se, \{maxime.bouton, jaeseong.jeong, ezeddin.al.hakim\}@ericsson.com}}
\begin{document}
\maketitle
\begingroup\renewcommand\thefootnote{\textsection}
\footnotetext{Equal contribution.}
\begin{abstract} 
6G will move mobile networks towards increasing levels of complexity. To deal with this complexity, optimization of network parameters is key to ensure high performance and timely adaptivity to dynamic network environments. The optimization of the antenna tilt provides a practical and cost-efficient method to improve coverage and capacity in the network. Previous methods based on Reinforcement Learning (RL) have shown effectiveness for tilt optimization by learning adaptive policies outperforming traditional tilt optimization methods. However, most existing RL methods are based on single-cell features representation, which fails to fully characterize the agent state, resulting in suboptimal performance. Also, most of such methods lack scalability and generalization ability due to state-action explosion. In this paper, we propose a Graph Attention $Q$-learning (GAQ) algorithm for tilt optimization. GAQ relies on a graph attention mechanism to select relevant neighbors information, improving the agent state representation, and updates the tilt control policy based on a history of observations using a Deep $Q$-Network (DQN). We show that GAQ efficiently captures important network information and outperforms 
baselines with local information by a large margin. In addition, we demonstrate its ability to generalize to network deployments of different sizes and density.
\end{abstract}
\section{Introduction}
\label{sec:intro}
With the growing complexity and densification of modern mobile networks, optimizing network parameters is becoming a challenging task for mobile network operators. Sixth-generation (6G) networks are expected to have a significant degree of heterogeneity with complex network topologies and high-dimensional observations. These features represent a serious challenge for network parameter optimization. With large amount of data in 6G intelligent network platform, data-driven network optimization will play a key role in enhancing network performance by automatically adapting network parameters to the dynamics of the environment. Network parameters, such as antenna height, transmit power, sector shape, vertical antenna tilt angle, etc., have a significant impact on the overall network performance. Specifically, the vertical antenna tilt plays a key role in determining the coverage and capacity experienced by User Equipments (UEs) in the network. The remote electrical control of the antenna tilt is an efficient and cost-effective method to improve Key Performance Indicators (KPIs) reflecting the Quality of Service (QoS) experienced by UEs. The methods presented in this work show how to control this parameter using Reinforcement Learning (RL) but can also be applied to other base station parameters for next generation networks.
\begin{figure}
    \centering
    \resizebox{0.48\textwidth}{!}{%
    \input{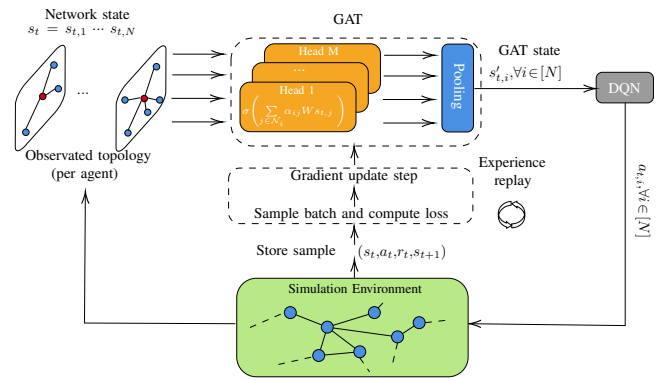}
    }
    \caption{Schematic representation of the GAQ algorithm.}
    \label{fig:gaq_figure}
    \vspace{-0.75cm}
\end{figure}

RL provides a powerful framework to learn dynamic tilt control strategies. Through repeated interactions with an environment, the agent adapts the tilt angle configuration based on network observations and a reward function specifying the desired agent behaviour. Most of the previous RL approaches to tilt optimization mainly relied on a set of KPIs characterizing the features of each cell individually~\cite{Berger14, Dandanov17}. These features were then used as state representation for learning algorithms such as Deep $Q$-Network (DQN)~\cite{mnih2013atari} to learn an optimal antenna tilt policy. In these methods, the agents do not consider explicitly the impact of neighbor antennas on its performance. Consequently, each antenna can only rely on its own state feature vector to learn the tilt control policy. Such representation may fail to capture mutual relations between neighboring antennas in the network, especially in complex and dense networks expected in 6G, and leads to suboptimal policies. \citet{whiteson2018qmix} has explored joint formulation between agent's action-value and global reward, which failed to learn the variational effect of agent's action affecting each of its neighbors \cite{rashid2020weighted}.

To capture these relations, Graph Attention Networks (GATs) \cite{velickovic2018gat} provide an efficient representation of the features of the \textit{target cell} (i.e., the cell considered for the antenna tilt update) and its neighbors, resulting in a richer state information. The GAT model proposed by \citet{brody2021attentive} exhibits better performance 
in capturing and enhancing pair-wise nodes importance with respect to classical Graph-Neural Networks (GNNs) based on graph convolution \cite{kipf2016semi, hamilton2017inductive}. In this work, we propose the Graph Attention $Q$-Network (GAQ) algorithm for tilt optimization, (illustrated in Fig. \ref{fig:gaq_figure}) which efficiently leverages neighbors information for representing the state of an RL agent.

Our proposed model introduces neighboring antenna information into the agent's observation space, without burdening the inter-cell signaling with more distant cell information. Our algorithm combines GAT with DQN, and demonstrates that it outperforms DQN, and other multi-agent RL methods with joint action-value \cite{whiteson2018qmix}, in a realistic simulated network environment. We demonstrate the ability of the learned model to generalize to large and denser network as well as varying number of neighbors without retraining. We also compare our approach against a competitive baseline using an Artificial Neural Network (ANN) to process neighbor information. In addition, we show the geographical distribution of the attention feature strengths across the network environment providing an intuitive interpretation of the relations between cells.

The remainder of the paper is structured as follows. In \cref{sec:related_work} we provide a review of related works on the tilt optimization along with works combining RL with GNNs. \Cref{sec:method} provides a comprehensive introduction to RL, DQN, and GATs, the main building blocks of our algorithm. In \cref{sec:gaq_ret}, we show how to formulate the tilt optimization problem in the RL setting and apply the GAQ algorithm for the remote control of antenna tilts. In \cref{sec:exp}, we describe the experimental setting and we demonstrate the empirical performance of our algorithm. Finally, in \cref{sec:conclusion}, we provide conclusive remarks and potential future works.
\section{Related work}
\label{sec:related_work}
There is a large body of works addressing antenna tilt optimization using RL \cite{Islam12, Guo13, Eckhardt11, Razavi10}. These approaches mainly relied on handcrafted state representations to capture the influence of neighboring actions on the reward, i.e., the state and reward were modeled using KPIs capturing the target cells features and its neighbors. However, such  state representation may fail to capture interference phenomena in complex and highly stochastic networks expected in 6G. 

Other works studied the antenna tilt problem in the multi-agent RL setting. \citet {bouton2021coordinated} approached the problem using based coordination graphs and pairwise relations between agents. While promising, this approach requires one model for each edge of the graph and a fixed message passing algorithm. Our formulation allows to learn using node features only and naturally embeds the message passing procedure in the GAT layer. A closely related work \citet{Balevi19} used a mean-field RL approach with linear function approximation, in which the influence from the neighbors is aggregated before a joint action within the neighbors is executed. However, this approach does not distinguish which neighbor has the most influence. In contrast, we use GAT to augment agent features with neighbor information, while capturing the importance of the neighbors through the attention mechanism, and use such representation as a state for DQN. 

Various works explored the combination of multi-agent RL and GNNs  \cite{jiang2020graph, liu2020multi, shao2021graph,chen2020gama}. The majority of these methods, focused on learning expressive representations of graph-structured data for the state of the agent. A major difference of our GAQ algorithm from the above works is the graph formulation. The works \cite{liu2020multi,jiang2020graph} are framed in the Relational Reinforcement Learning (RRL) setting, in which edges of the graph capture pair-wise relation between nodes, regardless of their adjacencies in the physical system. 
\cite{shao2021graph, liu2020multi, jiang2020graph} introduced attention mechanism in GNN for extracting agent state embedding.å While for all mentioned work above, the relation type between agent is predefined (e.g., collaborative or adversarial) and static over learning process.
While in antenna tilt optimization problem, the agent relation type is non-static (i.e.: cooperative or competitive) as state changes. 
Moreover, \citet{shao2021graph} apply graph attention layer on agent adjacencies, as well as distant ones, which is infeasible in tilt optimization problem.

\section{Background}
\label{sec:method}
\subsection{Reinforcement Learning}
RL is a decision-making framework in which an agent learns a control policy to maximize the cumulative reward through repeated interactions with an unknown environment. An RL problem can be formally modeled as a Markov Decision Process (MDP), defined by the tuple $(\mathcal{S}, \mathcal{A}, T, r, \gamma)$, where $\mathcal{S}$ is the state space, $\mathcal{A}$ the action space, $T$ is the unknown transition model, $r$ is the reward function, and $\gamma\in[0,1)$ is the discount factor. 
At time $t$, in a given state $s_t$, the agent takes an action $a_t$ and the environment transitions to a new state $s_{t+1}$. After every transition, the agent receives a reward $r_t\in\mathbb{R}$. The agent aims at finding a policy $\pi: \mathcal{S}\rightarrow\mathcal{A}$ that maximizes the expected discounted cumulative reward. We denote by $Q(s_t,a_t)=\mathbb{E}[\sum_{\tau=0}^\infty\gamma^\tau r_\tau|s_0=s]$, the state-action value of policy $\pi$. When the state space is continuous and high dimensional, the state-action value function can be approximated using parametric functions such as ANNs. In such case the $Q$-value is parametrized by a weight vector $w$. Based on this function representation the DQN algorithm seek to derive an approximately optimal weight vector for the value function by minimizing the following loss:
$$
J(w) = \mathbb{E}_{s_{t+1}}[(r_t + \gamma \max_{a} Q(s_{t+1},a) - Q(s_t, a_t))^2],
$$
Given a training sample, or a batch of experience, $(s_t, a_t, r_t, s_{t+1})$, the weights are updated using gradient ascent on $J(w)$ as:
$$
    w \leftarrow \alpha(r_t + \gamma\max_{a}Q(s_{t+1},a) - Q(s_t, a_t))\nabla_{w} Q(s_t, a_t),
$$
where $\alpha$ is a learning rate. In our algorithm, we use the DQN method proposed by \citet{mnih2013atari}, with improvements on the use of a target network and prioritized experience replay.

\subsection{Graph Attention Network}
\label{sec:gat}
This section presents the functioning principles of a graph attention network. For simplicity, we describe the functioning of a single GAT layer. A graph attention network is then formed by stacking up multiple GAT layers sequentially. The setup described in this section closely follows the work of \citet{velickovic2018gat}. We assume the existence of an underlying graph structure $\mathcal{G} = (\mathcal{V},\mathcal{E})$, where $\mathcal{V} = [N]$ and $\mathcal{E} \subseteq \mathcal{V}\times \mathcal{V}$ are the set of nodes and edges, respectively, and $N$ is the number of nodes. 

For a node $i$ in a graph, we denote by $\mathcal{N}_i$ the set of neighbors of $i$, and define $\mathcal{N}_{+i} = \mathcal{N}_i\cup \{i\}$. The input to the GAT layer is defined as a set of node features, $h = \{h_1,\dots, h_N\}$, where $h_i \in \mathbb{R}^d$, and $d$ is the number of node features. The layer produces, as output, a new set of processed node features $h' = \{h'_1,\dots, h'_N\}$, where $h'_i\in\mathbb{R}^{d'}$. The GAT layer first applies a linear transformation to map the input features into a higher dimensional space having dimension $d'\ge d$. To this end, the GAT uses a linear transformation parametrized by a weight matrix $W \in \mathbb{R}^{d'\times d}$. Then, the attention mechanism is applied to a node pair $(i,j)$, for $i\in \mathcal{V}$, and $j \in\mathcal{N}_i$, using a feed-forward ANN layer with a Leaky-ReLu activation function followed by a softmax layer to normalize the coefficients. The above steps can be summarized by the following equation:
\vspace{0.1cm}
$$
\alpha_{ij} = 
        \frac{\exp(\mathrm{LeakyReLU}(w_a^\top {[W h_{i} \Vert W h_{j}]}))}
    {\sum_{k\in \mathcal{N}_{+i}}
    \exp(\mathrm{LeakyReLU}(w_a^\top [W h_{i} \Vert W h_{k}]))},
$$
\vspace{0.1cm}
where $w_a\in \mathbb{R}^{2d'}$ is a learnable parameter vector and $[x\|y]$ represents the (column-wise) concatenation of the vectors $x$ and $y$. The normalized attention coefficient $\alpha_{ij}$ indicates the importance of node $j$ to node $i$. Finally, the normalized attention coefficients are used to compute a linear combination of the features, to serve as the final output features as follows:
\begin{equation}
\label{eq:gat_feat}
    h'_i = \sigma\left ( \sum_{j \in \mathcal{N}_{+i}} \alpha_{ij} W h_j\right),
\end{equation}
where $\sigma$ represents an activation function. It is reported in  \cite{velickovic2018gat} that, in order to stabilize the learning process, it is useful to extend the attention mechanism to employ multiple \textit{heads}, where one head represents the set of transformations leading to Eq. \eqref{eq:gat_feat}. Specifically, $M$ independent attention mechanisms are used, resulting in the output feature for node $i$:
\vspace{0.1cm}
\begin{equation}
\label{eq:gat_feat_heads}
h_{i}^{\prime}=\sum_{m\in[M]} \sum_{j \in \mathcal{N}_{+i}} \alpha_{i j}^{m} W^{m} h_{j},
\end{equation}
where the superscript $m$ denotes the $m$-th head of the attention coefficient $\alpha_{ij}^m \in\mathbb{R}$ and the weight matrix $W^m \in\mathbb{R}^{d'\times d}$.
\section{GAQ for Antenna Tilt Optimization}
\label{sec:gaq_ret}
In this section, we present the GAQ algorithm, we  formulate the antenna tilt optimization problem in the RL framework, and we show how to apply the GAQ algorithm for tilt optimization.
\subsection{Algorithm}
The GAQ algorithm combines DQN with GAT layers to enable sequential decision making on graph-structured inputs.
Specifically, the GAT builds an efficient state representation by aggregating neighbors features and learning neighboring nodes influence automatically. Such state representation is then fed to the DQN model to learn state-action values of the graph embedding and produce an optimal policy. 
The GAT acts on the state $s_{t,i}$ of each node $i \in \mathcal{V}$ at time step $t\ge 1$, by producing an augmented state representation. Specifically, based on the observed states $s_{t, \mathsmaller{\mathcal{N}_{+i}}} = \{s_{t,j}\}_{j \in \mathcal{N}_{+i}}$, the algorithm builds the GAT state $s'_{t,i}$ by applying the steps described in Sec. \ref{sec:gat} on the observed states $s_{t, \mathsmaller{\mathcal{N}_{+i}}}$. Note that the GAT state $s'_{t,i}$ has an implicit dependence on the GAT weights $w_a$ and $W$. This state is then processed by DQN with an $\varepsilon$-greedy exploration strategy and experience replay \cite{mnih2013atari}. At time $t$, and for node $i$, the agent selects a $\varepsilon$-greedy action as:
\begin{equation}
    \label{eq:sampling_rule}
    a_{t,i} = \begin{cases}
     a \sim \mathcal{U}(\mathcal{A}) &\text{w.p. } 1-\varepsilon \\
     \arg\max_{a} Q(s'_{t,i},a) & \text{w.p. } \varepsilon
    \end{cases}
\end{equation}
where $\mathcal{U}(\mathcal{A})$ denotes a discrete uniform random probability distribution over the action space $\mathcal{A}$. The replay buffer contains tuple of experience from all the cells and they are sampled using prioritized replay. When sampling there is no distinction of the cell as we are sharing the same model for all cells. When the DQN agent selects an action on the current node using this augmented state, it will consider the impact of the action on the current node’s and the neighboring nodes.  The algorithm pseudo-code is presented in \cref{alg:GAQ}. 
\begin{algorithm}[ht]
\SetAlgoLined
\textbf{Initialize}: DQN weights $w$, GAT weights $w_a$ and $W$, replay buffer $\mathcal{D} = \emptyset$\;
\For{$t = 1,\dots,T$}{
    \For{$i = 1 \dots,N$}{
        1. Observe states $s_{t, \mathsmaller{\mathcal{N}_{+i}}}$, \\
        2. Compute $s'_{t,i}$ as in \eqref{eq:gat_feat_heads}, \\
        3. Select action $a_{t,i}$ as in \eqref{eq:sampling_rule}, transition to state $s_{t+1, i}$, receive reward $r_{t,i}$, \\
        4. Append $(s_{t, \mathsmaller{\mathcal{N}_{+i}}}, a_{t,i}, r_{t,i}, s_{t+1, \mathsmaller{\mathcal{N}_{+i}}}) \to \mathcal{D}$, \\
     }
    5. Sample batch $(s_{t,i}, a_{t,i}, r_{t,i}, s_{t+1, i})$ from $\mathcal{D}$,\\ 
    6. Assign $y \gets r + \gamma \max_{a} (Q(s'_{t+1, i}, a))$,\\
    7. Perform a gradient update step on $\mathcal{L}(W,w_a,w) = [y - Q(s'_{t,i}, a_{t,i})]^2$. 
 }
\caption{GAQ}\label{alg:GAQ}
\end{algorithm}
\subsection{Antenna Tilt Optimization: Problem Formulation}
We now discuss how to build the relation graph $\mathcal{G}$ and present the RL problem formulation for the tilt optimization problem. The environment consists of a simulated mobile network in an urban environment. The environment is depicted in \cref{fig:environment}, and its parameters are presented in \cref{table:sim}. Each network cell is associated to a learning agent. From the point of view of a cell, the environment is formally defined as an MDP. Our algorithm searches for a policy $\pi(s_{t,i})$, for all cells $i \in\mathcal{C}$. The policy is learned across data samples from every cell in the network. Learning a policy in this way has the advantage of simplicity and is easier to learn as we can exploit the data from all sectors. 

\noindent \textbf{Graph construction.} The network environment can be represented as a graph. Nodes in the graph represent cells in the networks and edges represent relations between cells. We build the relation graph $\mathcal{G}$ based on the geographical distances and interference patterns of the antennas in the environment. The edges of the graph are used by the GAT layer to compute the state representation. The attention mechanism is expected to automatically identify relevant neighbors. The sparsity of the graph increases with the intersite distance. By varying this distance during training, we expose the agent to different edge relations.
\Cref{fig:environment} illustrates the generated graph for a hexagonal scenario with intersite distance of \SI{1}{\kilo\meter}. 
Each base station is composed of three cells and there is always an edge between cells belonging to the same base stations. In real networks,
edges in the graph can be discovered by a neighbor identification algorithm specified in 3GPP and cell handover logs~\cite{aliu2013}. Analyzing the influence of the graph structure is left as future work. 
\begin{figure}
    \centering
    \input{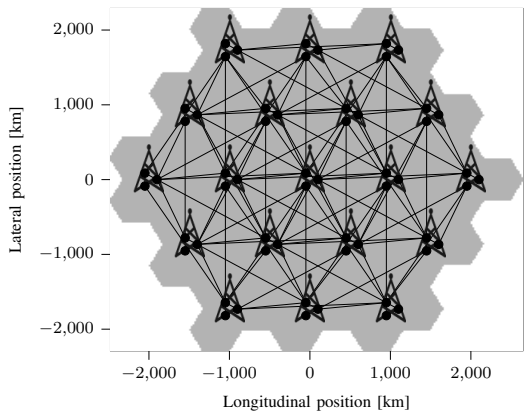}
    \vspace{-0.1cm}
    \caption{Network environment for an inter-site distance of \SI{1}{\kilo\meter}.} 
    \label{fig:environment}
    \vspace{-0.6cm}
\end{figure}

\noindent \textbf{MDP formulation.} We now specify the MDP formulation of the antenna tilt optimization problem.
\vspace{-0.1cm}
\begin{enumerate}
    \item \textit{State} at time $t$ and for cell $i$: $s_{t,i}$, a set of KPIs describing the state of a given cell. This includes the following features: \begin{inparaenum}[\itshape (i)\upshape] \item antenna position $ (x_i, y_i)$, \item normalized antenna direction  $(x_i^\Delta,y_i^\Delta )$, \item the $10$, $50$ and $90$ percentiles of the user's Signal-to-Interference-plus-Noise Ratio (SINR), and \item the antenna tilt angle $\theta_{t,i}$. \end{inparaenum}
    \item \textit{Action} at time $t$ for cell $i$: the agent selects a tilt angle $a_{t,i} \in [0^{\circ},15^{\circ}]$. At the next time step $t+1$, the tilt angle for sector $i$ is $\theta_{t+1,i} = a_{t,i}$.
    \item \textit{Reward} at time $t$ and for cell $i$:  $r_{t,i} = \frac{1}{|\mathcal{N}_{+i}|} \sum_{j \in \mathcal{N}_{+i}} \text{SINR}(t,j)$, where $\text{SINR}(t,j)$ denotes the average SINR of users in cell $j$. 
\end{enumerate}

\section{Experimental Setup \& Results}
\vspace{-0.15cm}
\label{sec:exp}
\subsection{Experimental setup}
\vspace{-0.1cm}
\label{subsec:sim}
\noindent \textbf{Simulator.} We run our experiments in a proprietary mobile network simulation environment. The simulation is executed on an urban network consisting of $\mathcal{B}$ BSs, $\mathcal{C}$ cells, and $\mathcal{U}$ users randomly positioned in the environment. 
The network environment is shown in Fig. \ref{fig:environment}, and the simulation parameters are reported in Tab. \ref{table:sim}. 
\begin{table}[h!]
  \caption{Simulation Environment Parameters}\label{table:sim}
  \vspace{-0.2cm}
  \centering 
  \begin{threeparttable}
    \begin{tabular}{p{3cm}p{1cm}p{3cm}}
    \midrule
    \textbf{Parameter} & \textbf{Symbol} & \textbf{Value} \\
     \midrule\midrule
Number of cells  & $\mathcal{C}$ & \num{57} training, \num{111} evaluation\\
\midrule
Number of UEs  & $\mathcal{U}$& $1000$\\ \midrule
Number of BSs  & $\mathcal{B}$& \num{19} training, \num{37} evaluation\\ \midrule
Number of neighbors & $|{\mathcal{N}(i)}|$ & $5$\\\midrule
Inter-site distance & $d$ & [300,1500] \SI{}{\meter}\\
\midrule
Antenna height & ${h}$ & \SI{32}{\meter}\\
\midrule
Frequency & $f$ & \SI{2.1}{\GHz}\\
\midrule
Traffic volume & $\tau$ & \SI{1}{Mbps}\\
\bottomrule
\end{tabular}
\vspace{-0.4cm}
\end{threeparttable}
\end{table}
Based on the user positions, antennas positions and directions, the simulator computes the path loss in the network environment using the Okomura-Hata propagation model \cite{Rappaport96}, and returns the SINRs values by conducting user association and resource allocation in a full-buffer traffic demand scenario. To promote robustness of the learned policy, we reset the environment parameters every \num{20} steps by selecting them (uniformly) at  random from the ranges specified in  \cref{table:sim}. In particular, the parameter we vary is the inter-site distance $d$.  

\noindent \textbf{Baseline policies.} In our experiments, we compare the proposed GAQ policy, with a DQN baseline \cite{mnih2013atari}, which run independently in each cell, a QMIX\cite{whiteson2018qmix} baseline, and a Neighbor-DQN (N-DQN) policy, in which the state of the agent includes observations from the cell of interest and the neighbors. All approaches learn a shared model for all the cells and can leverage data from the whole network. We conduct $3$ random and independent training and report the performance in terms of mean and standard deviation.  

\noindent \textbf{Training parameters.} We train the policies for \num{20 000} time steps, with a learning rate of \num{1e-3} and a replay buffer of \num{10 000} data points. The ANN parametrizing the $Q$-function is a multi-layer perceptron with \num{2} layers of \num{32} nodes,  with a ReLU activation function. The GAT architecture contains \num{2} layers and \num{6} heads. The exploration schedule for  $\varepsilon$-greedy decays from $1$ to $1\times 10^{-2}$, for $t \in [0,T/2]$ time steps, and stays constant for $t \in [T/2,T]$, where $T$ is the number of training steps. All training policies follow a DQN-like off-policy training with double $Q$-learning and prioritized experience replay~\cite{Hasselt2016}.

\vspace{-0.1cm}
\subsection{Results \& Discussion}
\vspace{-0.1cm}
We report, in Fig. \ref{fig:results}, the learning curves for the GAQ, DQN, QMIX and N-DQN policies. The curves represent the average reward over cells: $\Bar{r}_t = \frac{1}{N} \sum_{i \in [N]} r_{t,i}$. Specifically, the continuous line and the shaded area represent the mean and standard deviation of the average reward performance, respectively, across $K = 3$ independent runs. We show that GAQ outperforms DQN and QMIX, and approximately matches the performance of N-DQN. In this scenario, the N-DQN algorithm serves as an oracle as we consider a fixed number of neighbors, it learns to represent the best way to process those neighbors through an ANN. This shows that the GAT layer has significant expressivity to capture complex relations between a cell and its neighbors. The drawback of N-DQN is that it cannot adapt to larger number of neighbors without an expensive retraining. 
\begin{figure}[h!]
    \vspace{-0.2cm}
    \centering
    \includegraphics[width = 0.55\columnwidth]{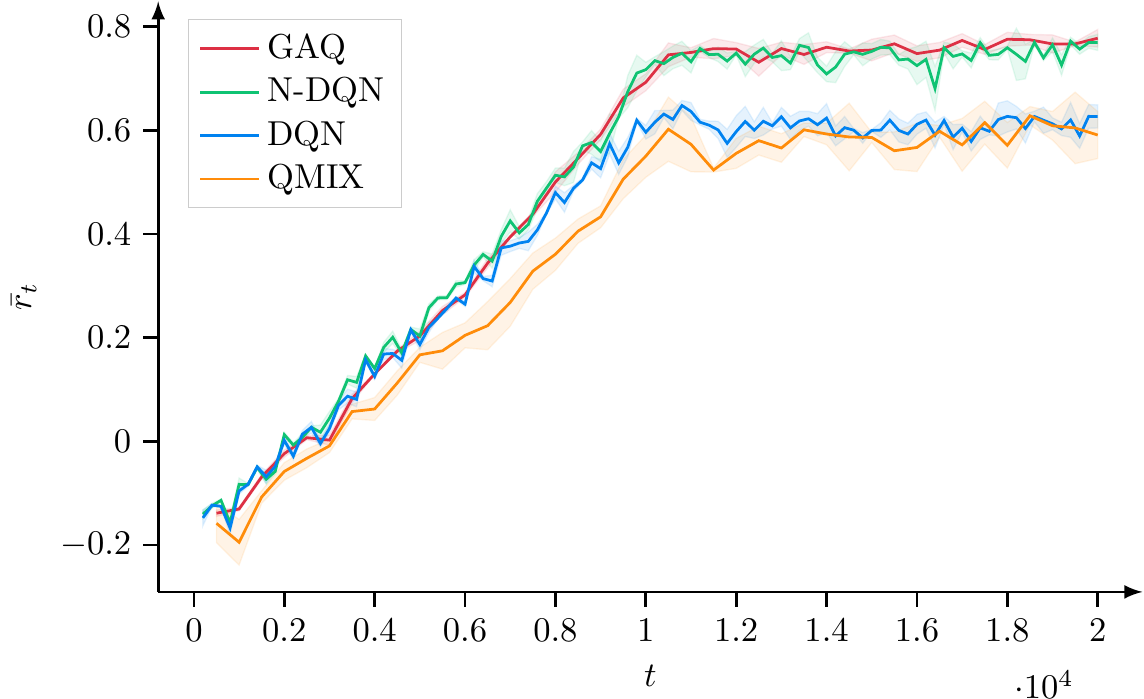}
     \vspace{-0.2cm}
    \caption{Average training reward.}
    \vspace{-0.2cm}
    \label{fig:results}
\end{figure}

As stated in \cite{Knyazev2019understanding}, GAT shows significant generalization ability, i.e., it is able to achieve good performances when tested on  environment different from the ones on which it is trained on. For this reason, we report, in Fig. \ref{fig:test_results}, the generalization performance of the GAQ trained on $5$ neighbors (GAQ-$5$) environment and test it on $10$ and $20$ neighbors (GAQ-$10$ and GAQ-$20$, respectively). Specifically, we show the CDFs of the average reward during the evaluation across
$3$ independent runs and $100$ episodes. 
\begin{figure}[h!]
 \vspace{-0.2cm}
    \centering
    \includegraphics[width = 0.6\columnwidth]{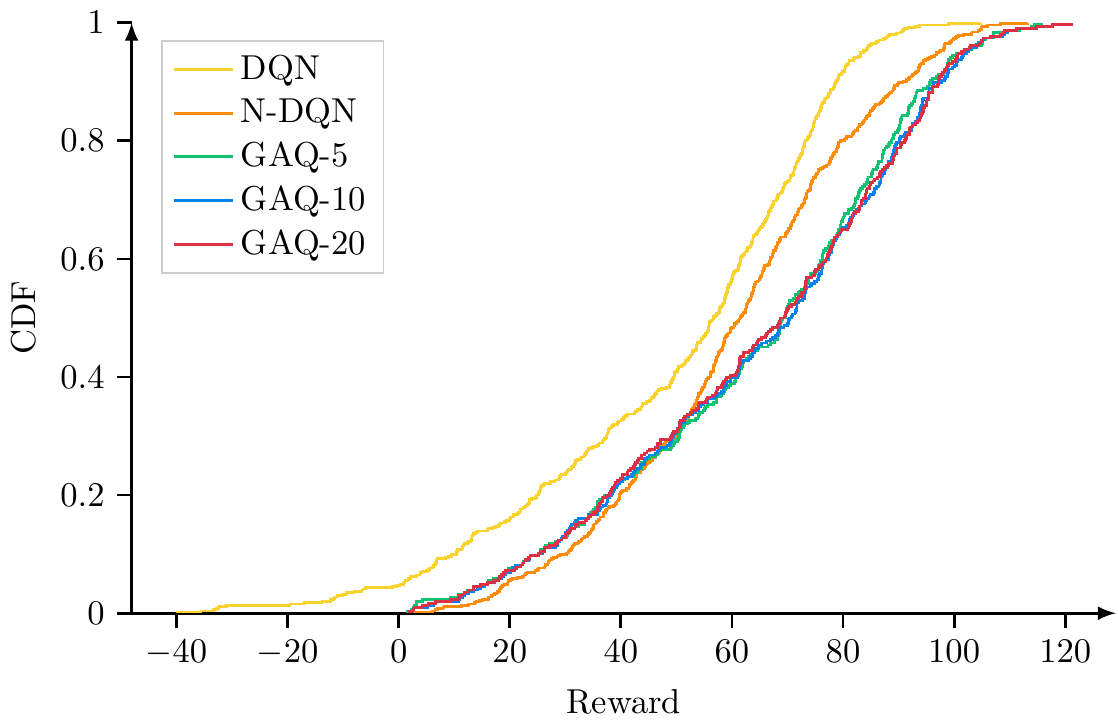}
     \vspace{-0.2cm}
    \caption{Average test CDFs for different neighbor set size.}
     \vspace{-0.2cm}
    \label{fig:test_results}
\end{figure}

Fig. \ref{fig:test_results} shows that similar performances are achieved by GAQ-$5$, GAQ-$10$, and GAQ-$20$ with a slight improvement when including more neighbors. This indicates good generalization performance of the GAQ trained on $5$ neighbors on an increased number of neighbors, and suggests potential scalability to larger or denser network scenarios.

Finally, in Fig. \ref{fig:AxcelAttn}, we show the strength of the attention coefficients across the mobile network graph, in terms of the average over the different attention heads. We may observe that the attention coefficients for each node show a similar pattern: the attention strength between antennas belonging to the same BS is generally higher, while the one between antennas in two different BS is high only when their orientation matches. 
\begin{figure}[h!]
    \vspace{-0.45cm}
    \centering
    \includegraphics[width = 0.8\columnwidth]{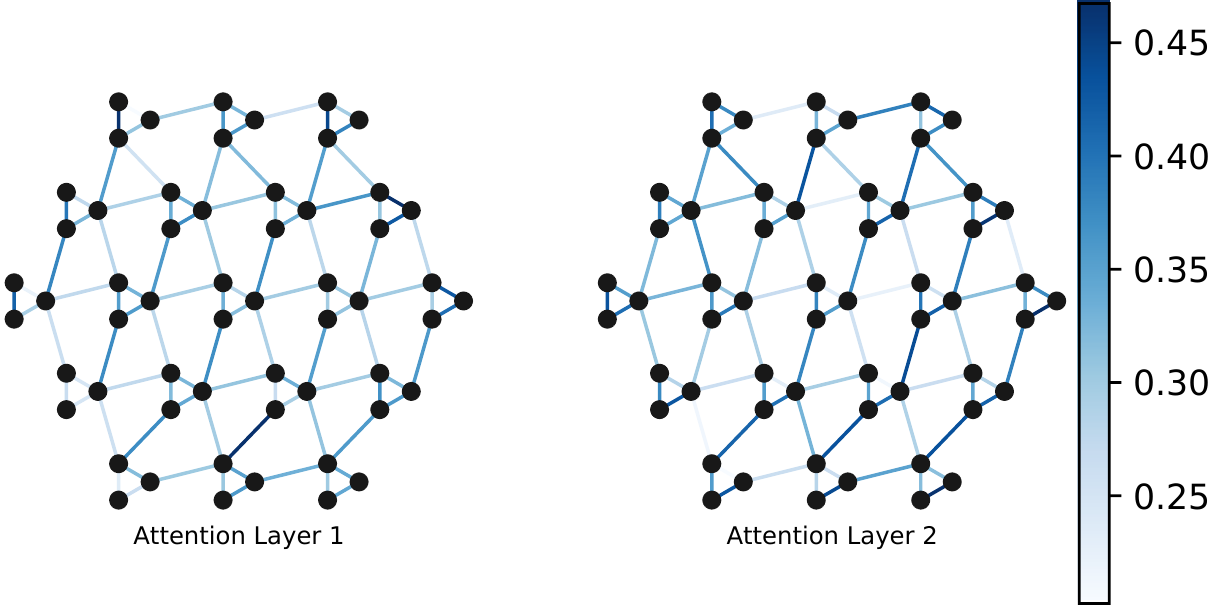}
    \caption{Attention coefficients strength across the network.}
    \label{fig:AxcelAttn}
    \vspace{-0.2cm}
\end{figure}
\vspace{-0.2cm}
\section{Conclusions}
\vspace{-0.15cm}
\label{sec:conclusion}
In this paper, we have introduced GAQ, an effective learning algorithm combining GAT and DQN, and applied our algorithm to the antenna tilt optimization problem. Extensive experimental results on a realistic mobile network environment demonstrated the empirical effectiveness of the proposed methodology: the GAQ policy outperforms DQN and QMIX and achieves similar performance to N-DQN in our environment. The GAQ policy also generalizes to denser networks, making this method particularly appealing to 6G networks. It is worth noting that the GAQ method may be easily applied to other network optimization use cases which share a similar graph structure (e.g., down-link power control, radiation pattern optimization, beam management in MIMO systems, etc.). 

\section*{Acknowledgement}
\vspace{-0.15cm}
This work was partially supported by the Wallenberg AI, Autonomous Systems and Software Program (WASP) funded by the Knut and Alice Wallenberg Foundation.
\addtolength{\textheight}{-6cm}   
\vspace{-0.15cm}
\printbibliography

\end{document}